\documentclass[letterpaper]{article} % DO NOT CHANGE THIS
\usepackage[draft]{aaai25}  % DO NOT CHANGE THIS
\usepackage{times}  % DO NOT CHANGE THIS
\usepackage{helvet}  % DO NOT CHANGE THIS
\usepackage{courier}  % DO NOT CHANGE THIS
\usepackage[hyphens]{url}  % DO NOT CHANGE THIS
\usepackage{graphicx} % DO NOT CHANGE THIS
\usepackage{natbib}  % DO NOT CHANGE THIS AND DO NOT ADD ANY OPTIONS TO IT
\usepackage{caption} % DO NOT CHANGE THIS AND DO NOT ADD ANY OPTIONS TO IT
\usepackage{algorithm}
\usepackage{algorithmic}
\usepackage{multirow}
\usepackage{booktabs}
\usepackage{subcaption}
\usepackage{caption}
\usepackage{amssymb}
\usepackage[para, hang, flushmargin]{footmisc}
\usepackage{newfloat}
\usepackage{listings}
\usepackage{amsmath}
\usepackage{mathtools}
\usepackage{breqn}
\DeclareCaptionStyle{ruled}{labelfont=normalfont,labelsep=colon,strut=off} % DO NOT CHANGE THIS
\floatstyle{ruled}
\newfloat{listing}{tb}{lst}{}
\floatname{listing}{Listing}
\title{Modulated Intervention Preference Optimization (MIPO):\\ Keep the Easy, Refine the Difficult}
\author{
    Cheolhun Jang\\
}
\affiliations{
    NC Research\\
    9405hun@ncsoft.com
}

\usepackage{bibentry}
\begin{document}

\maketitle

\begin{abstract}

Preference optimization methods typically begin training with a well-trained SFT model as a reference model. In RLHF and DPO, a regularization term is used during the preference optimization process to prevent the policy model from deviating too far from the reference model's distribution, thereby avoiding the generation of anomalous responses.
When the reference model is already well-aligned with the given data or only requires slight adjustments, this approach can produce a well-aligned model. However, if the reference model is not aligned with the given data and requires significant deviation from its current state, a regularization term may actually hinder the model alignment.
In this study, we propose \textbf{Modulated Intervention Preference Optimization (MIPO)} to address this issue. MIPO modulates the degree of intervention from the reference model based on how well the given data is aligned with it. If the data is well-aligned, the intervention is increased to prevent the policy model from diverging significantly from reference model. Conversely, if the alignment is poor, the interference is reduced to facilitate more extensive training. We compare the performance of MIPO and DPO using Mistral-7B and Llama3-8B in Alpaca Eval 2.0 and MT-Bench. The experimental results demonstrate that MIPO consistently outperforms DPO across various evaluation scenarios.

\end{abstract}

% In MIPO, the degree to which the reference model is aligned with a given preference pair is measured by using average log likelihood of responses.
% Based on this, if the reference model is well-aligned to the given data, MIPO objective is designed to ensure that the policy model does not significantly diverge from the reference model's distribution. Conversely, if the model is poorly aligned to given the preference data, the MIPO objective is designed to allow the policy model to depart from the reference model and find more optimal weights. 


\section{Introduction}

As the performance of Large Language Models (LLMs) trained with a large amount of data has been attracting attention, methods \cite{plm, plm2,plm3} for training them have been actively studied. The commonly used LLM training pipeline is to pretrain LLM using a large amount of data, and then use the instruction-tuning method \cite{instruction-tuning4} to allow LLM to follow the human-provided instruction. 

However, it is difficult to train LLM to produce the desired output (helpful, harmless) or to prevent LLM from producing the output that LLM should not produce \cite{safety}. Therefore, LLM alignment methods employing human feedback have started to gain significant attention.

Among these methods, Reinforcement Learning from Human Feedback (RLHF) \cite{rlhf-ref1, rlhf-ref2} received significant attention. Models trained with RLHF are well-aligned with human feedback, demonstrating reliable performance as a result \cite{rlhf-helpful, rlhf-helpful2, rlhf-safe}. However, the RLHF approach involves a complex training process, including the training of a reward model, which has posed significant challenges in the implementation and training \cite{rlhf-problem1, rlhf-problem2}.

\begin{figure}[t]
  \centering
  \includegraphics[scale=0.53]{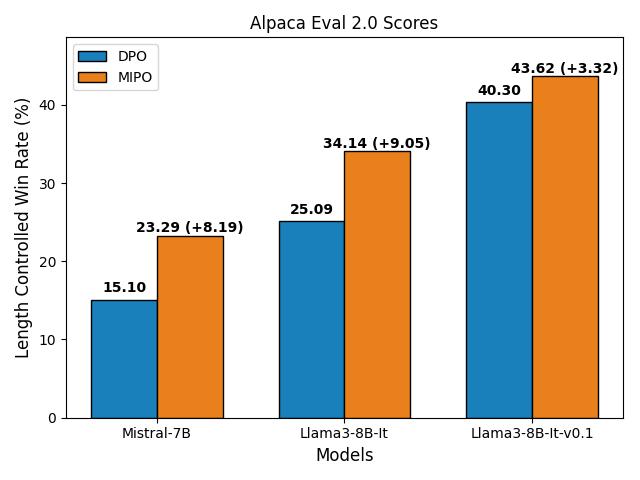}
  \caption{Alpacaeval 2.0 scores for MIPO and DPO implementations on Mistral-7B-Base and Llama-8B-Instruct. v0.1 is a model trained on different dataset.}
  \label{fig:Alpacaeval}
\end{figure}

Direct Preference Optimization (DPO) \cite{dpo} is a method designed to overcome these limitations. In DPO, the optimization problem of RLHF is modified to eliminate the reward model and train only the policy model. This makes it easier to train DPO compared to RLHF, and DPO also effectively learned human preferences, demonstrating strong performance.

In DPO and RLHF, the policy model is trained to align with the instance while ensuring its distribution does not move significantly away from the reference model's distribution to prevent it from generating anomalous responses (ex. hallucinations).
Therefore, if the reference model is moderately aligned with the given preference pair, it could be possible to train a well-aligned policy model for the given data without significantly diverging from the reference model's distribution.
However, if the reference model is not aligned with the given preference pair, it will be difficult for the policy model to align with the data through minor adjustments, without moving far from the reference model's distribution.
Therefore, it is crucial to adjust the training objective based on how well the reference model is aligned.

\begin{figure*}[t]
  \centering
  \includegraphics[width=\textwidth,scale=0.4]{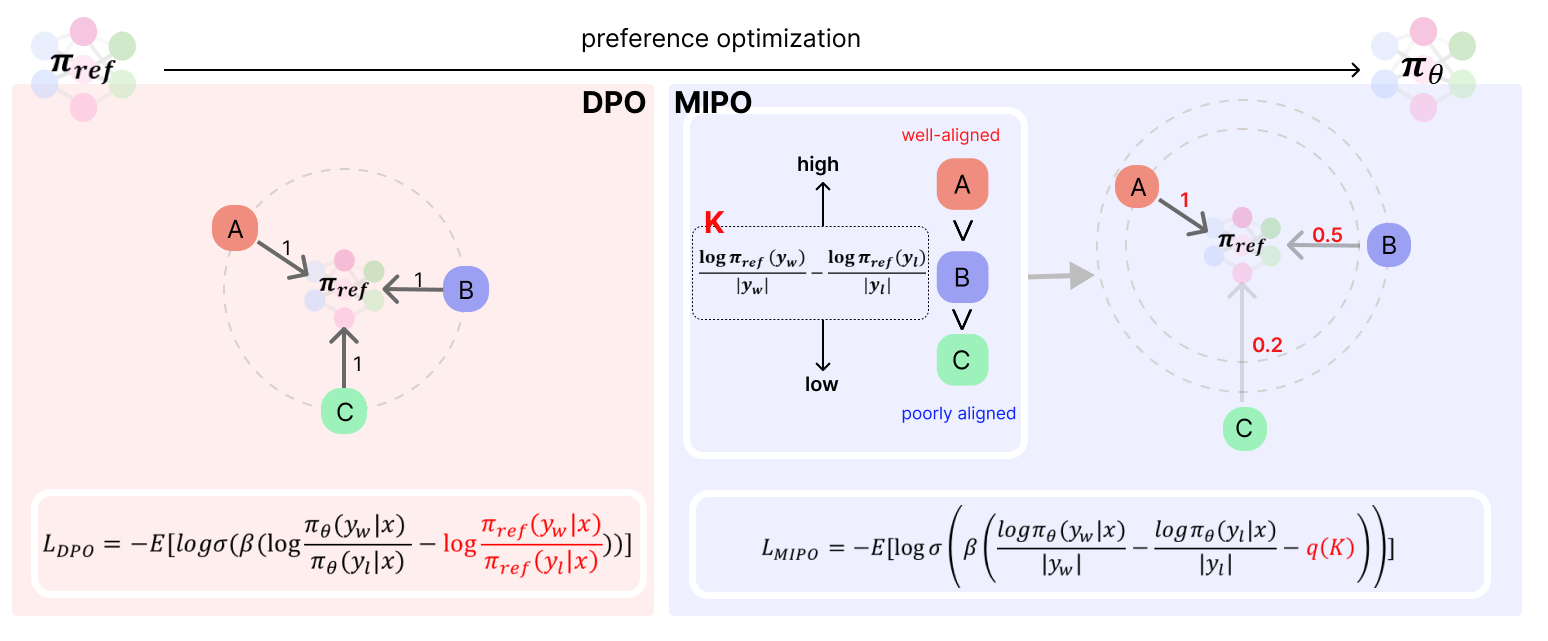}
  \caption{\textbf{Optimization process of MIPO}. In DPO, the objective utilizes a consistent regularization term (red part in DPO objective) for the reference model across all instances (\textit{A}, \textit{B}, \textit{C} in Figure), regardless of the degree of alignment of each instance. However, in MIPO, the alignment of each instance with the reference model is first assessed by using the difference in average log likelihood. Based on this value, $K$, the extent to which the reference model will intervene in the learning process is determined and subsequently reflected in the MIPO objective.}

  \label{fig:MIPO}
\end{figure*}

In this paper, we propose a preference optimization algorithm, \textbf{Modulated Intervention Preference Optimization} (MIPO), to address this issue. As seen in Figure \ref{fig:MIPO}, MIPO utilizes the average log likelihood to measure how well the reference model is aligned with the given preference pair. 
Through this value, the MIPO objective is configured to modulate the intervention of the reference model, allowing more extensive training on pairs that are judged to be poorly aligned with the reference model. 
We use Alpaca Eval 2.0 and MT-Bench to compare the performance of MIPO with DPO and other preference optimization methods. Across diverse experimental settings, MIPO consistently achieves outstanding performance.
To summarize, \textbf{MIPO has the following properties:}

\begin{itemize}
    \item \textbf{Modulate the intervention of the Reference Model}:
    MIPO is a novel approach that modulates the intervention of the reference model for each instance. It determines the extent of the reference model's intervention based on the degree of alignment. MIPO maintains performance on pairs where the reference model already well-aligned, while simultaneously achieving substantial performance gains on pairs where the reference model previously underperformed (Section \$\ref{avg log prob analysis}).

    \item \textbf{Outstanding Benchmark Performance}:
    We conduct experiments using Llama3-8B-Instruct \cite{llama3} and Mistral-7B-Base \cite{mistral} to verify the effectiveness of MIPO in various models.
    On Alpaca Eval 2.0, our proposed method consistently outperforms DPO.
    % A significant performance improvement is observed compared to DPO and other preference optimization methods in AlpacaEval 2.0. 
    As we can see in Figure \ref{fig:Alpacaeval}, in Llama3-8B-Instruct, it outperforms DPO by approximately 9 points (+36.07\%), and in Mistral-7B-Base, it outperforms about 8 points (+54.24\%). In most cases, MIPO achieves the best performance not only compared to DPO but also when compared to other methods. 
    On MT-Bench, MIPO also exhibits the best performance among the compared approaches (Section \$\ref{benchmark_results}).

    \item \textbf{Simple and Effective Training}:
    The high-performance model can be found in MIPO by tuning only the hyper-parameter $\beta$. Moreover, consistently outstanding performance is achieved within a specific range of $\beta$, independent of model architecture or dataset.
    Thus, unlike other methods that require extensive tuning, this approach allows for easy acquisition of a high-performance model with minimal tuning effort (Section \$\ref{6.2:training_stablity}).
    
\end{itemize}

\section{Related Works}

After being pretrained on a large amount of data \cite{plm} and fine-tuned \cite{instruction-tuning, instruction-tuning2}, LLMs have achieved notable performance across many tasks \cite{plm2,plm3, instruction-tuning3}.
However, LLMs that could generate responses that were even more helpful and harmless were needed, leading to the development of preference optimization methods \cite{rlhf-ref1, safety, safety2} that fine-tune LLMs more closely to human feedback.

RLHF \cite{rlhf-ref2, rlhf-ref3} is one such preference optimization method for LLM alignment. In RLHF, preference data is used to train a reward model, which is then utilized to optimize the policy model by Proximal Policy Optimization \cite{ppo}.
RLHF effectively aligns models with human feedback, resulting in good performance \cite{rlhf-helpful, rlhf-helpful2}. 
However, there are challenges, such as the difficulty of obtaining scored data, ensuring stable training, and the necessity of training a reward model \cite{rlhf-problem1, rlhf-problem2, rlhf-problem3}.

DPO is a preference optimization method that solves optimization problem of RLHF in a more easier and efficient manner. \cite{dpo} proposed DPO to eliminate the reward model in RLHF and train only the policy model with preference data. It is simple compared to RLHF, and the training phase is more stable. So it has become one of the widely used method for aligning language models.  
However, DPO also has its drawbacks like dependency on the reference model and issues with length exploitation \cite{dpo-problem, dpo-probelm-2, cpo}. Therefore, new model alignment methods such as KTO \cite{kto}, IPO \cite{ipo} and ORPO \cite{orpo} continue to emerge.

However, most methods including DPO does not take into account the differences in the degree of alignment of the reference model between preference pairs. As mentioned earlier, if the reference model is already well-aligned, only minimal training will be needed to achieve alignment. Conversely, if the reference model is completely misaligned, extensive training will be required. However, DPO does not account for these differences (Section \$\ref{3.3}).

To address this issue, we propose \textbf{MIPO}, which varies the learning weights among instances by modulates the degree of intervention from the reference model (Section \$\ref{4}).

\section{Background}

In this section, we will review the DPO in Section \$\ref{3.2}, and analyze the ineffective aspects of DPO in Section \$\ref{3.3}.

\subsection{Terminology}

\(D = \left \{{x^i,y_w^i,y_l^i} \right \}_{i=1}^{N}\) is for pairwise-preference dataset, where \(x^i\) is prompt and \(y_w^i\) is chosen (preferred) response and \(y_l^i\) is rejected (dis-preferred) response for that prompt. \( \pi_{ref}\) is reference model, initial LLM that we start training from. \( \pi_{\theta}\) is policy model, which is a model we train.

\subsection{DPO}
\label{3.2}
DPO employs the Bradley-Terry (BT) model \cite{bt} to represent the distribution of human preference. BT model represents human preference distribution for \(y_w\), \(y_l\) by the reward function as follows:

\begin{equation}
\label{eqn:1}
p(y_w > y_l | x) = \frac{exp(r(x, y_w))}{exp(r(x,y_w)) + exp(r(x,y_l))}
\end{equation}

DPO's reward function is reparameterized from the RLHF's objective as following equation.

\begin{equation}
\label{eqn:2}
r(x,y) = \beta log\frac{\pi_\theta(y|x)}{\pi_{ref}(y|x)} + \beta log Z(x)
\end{equation}

From equations \eqref{eqn:1} and \eqref{eqn:2}, we can formulate preference distribution by using \( \pi_{ref}\) and \( \pi_{\theta}\). Subsequently, the DPO objective is derived as expressed in \eqref{eqn:3}

\begin{dmath}
\label{eqn:3}
L_{DPO}(\pi_\theta; \pi_{ref}) = E_{(x, y_w, y_l)~\sim D }\left [- log \sigma (\beta log \frac{\pi_\theta(y_w|x)}{\pi_{ref}(y_w|x)} - \beta log\frac{\pi_\theta(y_l|x)}{\pi_{ref}(y_l|x)} ) \right ]
\end{dmath}

\subsection{Ineffective Aspects of DPO} \label{3.3}

\subsubsection{DPO does not consider how well the preference pairs are aligned.}

Looking at the reward of DPO in Eq \eqref{eqn:2} without \(Z(x)\). It can be seen that the reward is the difference between the log likelihood of the policy model and the log likelihood of the reference model. 
This implies that DPO allows for high rewards to be obtained solely by increasing the log likelihood of a response, without considering the degree to which the reference model already performs well on that response. Consequently, the training process proceeds without taking into account the extent to which the reference model is aligned with the give preference data.

For example, consider \textit{$pair_1$}, preference data where the reference model already well-aligned, and \textit{$pair_2$}, where it does not. Ideally, model will require to train slightly on \textit{$pair_1$} to maintain its current performance, while it will require substantial training for \textit{$pair_2$} compared to \textit{$pair_1$}. 

Let's assume that the policy model has been trained so that the log likelihood of the chosen response increases by $\alpha$ compared to the reference model, while the log likelihood of the rejected response remains unchanged in both pairs ($log\pi_\theta(y_w|x)-log\pi_{ref}(y_w|x) = \alpha,
log\pi_\theta(y_l|x)-log\pi_{ref}(y_l|x) = 0$).
In DPO, both pairs would yield the same loss by Eq \eqref{eqn:3}. This implies that the improvement in log likelihood for \textit{$pair_1$} and \textit{$pair_2$} holds equal significance in DPO.

Consequently, DPO trains the model without discriminating between instances of strong and weak alignment with the reference model.
This uniform approach can result in insufficient training for pairs where the reference model needs improvement and excessive training for pairs where preferences are already adequately captured. Therefore, this issue can negatively impact the performance of the trained model.

\section{Methodology} \label{4}

In this section, we explain why we use average log likelihood to determine how well reference model is aligned to data in Section \$\ref{4.1}. Then we introduce \textbf{Modulated Intervention Preference Optimization} (MIPO), an algorithm that adjusts the degree of intervention from the reference model based on the level of alignment in Section \$\ref{4.2}.

\subsection{Measuring the Alignment Degree} \label{4.1}
To solve the problem of DPO mentioned above Section \$\ref{3.3}, we first need to measure which pairs are well-aligned to reference model and which pairs are poorly aligned. 

In the context of preference learning, being "well-aligned" can be interpreted as the model being more likely to generate a chosen response $y_w$ than a rejected response $y_l$ for a given input $x$. However, using the difference in log likelihoods between chosen and rejected responses to measure alignment is not feasible, as log likelihood is highly sensitivity to response length. If the lengths of the chosen and rejected responses differ significantly, the longer response's log likelihood will be disproportionately lower, regardless of individual token probabilities.

Therefore, we decide to use of average log likelihood. It allows for a more fairer comparison of generation probabilities between chosen and rejected responses, mitigating the impact of length discrepancies. We have decided to use the difference in average log likelihood, $K$, as a metric to assess the alignment of the reference model with a given pair.

\begin{equation}
\label{eqn:4}
K =\frac{log\pi_{ref}(y_w|x)}{|y_w|} - \frac{log\pi_{ref}(y_l|x)}{|y_l|}
\end{equation}

We interpret a high $K$ value as indicative of strong alignment in the reference model, whereas a low $K$ value suggest insufficient alignment. Based on this assumption, we propose our objective as follows:

\subsection{Deriving the MIPO Objective} \label{4.2}

\begin{equation} \label{eqn:5}
\begin{split}
L_{MIPO}(\pi_\theta; \pi_{ref}) & = E_{(x, y_w, y_l)~\sim D } \\ &  - log \sigma (\beta
\underbrace{(\frac{log\pi_\theta(y_w|x)}{|y_w|} - \frac{log\pi_\theta(y_l|x)}{|y_l|})}_{\text{$f(\theta)$}}  \\
& - \beta \underbrace{ log(1 + e^K)}_{\text{\textit{$q(K)$}}} ) \\
\end{split}
\end{equation}

For the reasons mentioned above, the MIPO objective is designed to enhance the alignment of the policy model by using average log likelihood, $f(\theta)$. Additionally, it is adjusted based on the degree of alignment through $q(K)$, which acts as a modulator for the degree of intervention from the reference model. 

Let's examine the MIPO objective in two cases:
\subsubsection{When reference model is well aligned for a given pair}
It means $K$ is large enough. Then, $q(K)$ converges to $K$ and the objective of MIPO can be expressed as follows.

\begin{equation}
\label{eqn:MIPO}
\begin{split}
L_{MIPO} = - log \sigma (\beta  (\frac{log\pi_\theta(y_w|x)}{|y_w|} - \frac{log\pi_\theta(y_l|x)}{|y_l|}) \\
- \beta ( \frac{log\pi_{ref}(y_w|x)}{|y_w|} - \frac{log\pi_{ref}(y_l|x)}{|y_l|})).
\end{split}
\end{equation}

The objective is calculated based on the difference between the policy model's average log likelihood difference, $f(\theta)$, and this values of reference model, $K$. Therefore, as $f(\theta)$ diverges further from $K$, the loss decreases, preventing the policy model from significantly diverging from the reference model.

\subsubsection{When reference model is poorly aligned for a given pair}
It means $K$ is low. In this case, $q(K)$ approaches to $0$ and objective can be expressed as follows.

\begin{equation}
\begin{split}
L_{MIPO} = - log \sigma (\beta  (\frac{log\pi_\theta(y_w|x)}{|y_w|} - \frac{log\pi_\theta(y_l|x)}{|y_l|}))
\end{split}
\end{equation}
 
Since the MIPO objective does not include a term for the reference model, it only considers the $f(\theta)$ for alignment, focusing solely on increasing this value. When compared to the case where $q(K) = K$, it is clear that the MIPO loss significantly greater because $f(\theta)$ is less than $f(\theta) - K (\because K < 0)$. Consequently, the policy model can be trained while diverging further from the distribution of the reference model.

In summary, the MIPO assesses how well the reference model is aligned with the given instance through the metric $K$. This metric is then used to calculate $q(K)$, which determines the extent to which the reference model's influence on the policy model's learning. When $K$ is high, it indicates strong alignment with the given data. In this case, $q(K)$ takes on the value of $K$, thereby increasing the intervention of the reference model. Consequently, the policy model train without diverging significantly from the reference model. Conversely, if $K$ is low, $q(K)$ becomes zero, allowing the policy model to train without intervention from the reference model. 

More detailed explanations about objective are provided in the Section \$\ref{loss analysis} and gradient analysis can be found in Appendix A.

\begin{table*}[h]
\centering
\begin{tabular}{@{}cccccccccc@{}}
\toprule
\multirow{3}{*}{Method} & \multicolumn{3}{c}{Mistral-7B-Base}                 & \multicolumn{3}{c}{Llama3-8B-Instruct}         & \multicolumn{3}{c}{Llama3-8B-Instruct-v0.1}         \\ \cmidrule(l){2-4} \cmidrule(l){5-7} \cmidrule(l){8-10}  
                        & \multicolumn{2}{c}{Alpaca Eval 2.0} & MT-Bench      & \multicolumn{2}{c}{Alpaca Eval 2.0} & MT-Bench      & \multicolumn{2}{c}{Alpaca Eval 2.0} & MT-Bench      \\ \cmidrule(l){2-3} \cmidrule(l){4-4} \cmidrule(l){5-6} \cmidrule(l){7-7} \cmidrule(l){8-9} \cmidrule(l){10-10} 
                        & LC(\%)           & WR(\%)           & Avg. Score    & LC(\%)           & WR(\%)           & Avg. Score    & LC(\%)           & WR(\%)           & Avg. Score    \\ \midrule
ORPO                    & 14.7*            & 12.2*            & -             & -                & -                & -             & 28.5*            & 27.4*            & -             \\
KTO                     & 13.1*            & 9.1*             & -             & -                & -                & -             & 33.1*            & 31.8*            & -             \\
SimPO                   & 21.4*            & \textbf{20.8*}   & 7.05          & -                & -                & -             & \textbf{44.7*}            & 40.5*            & 7.72          \\
DPO                     & 15.1*            & 12.5*            & 7.01          & 25.09            & 21.18            & 7.95          & 40.3*            & 37.9*            & 7.79          \\
MIPO                     & \textbf{22.02}   & 17.50            & \textbf{7.12} & \textbf{34.14}   & \textbf{30.00}   & \textbf{7.97} & 43.62   & \textbf{40.74}   & \textbf{7.92} \\ \bottomrule
\end{tabular}
\caption{AlpacalEval 2.0 and MT-Bench scores for preference optimization methods in Mistral-7B, Llama3-8B. The v0.1 tag refers to a model trained using \textbf{Llama3 Ultrafeedback} data, and the others are all trained with \textbf{Binarized UltraFeedback}. Results denoted with (*) are sourced from \cite{simpo}.}
\label{table:overall}
\end{table*}

\section{Experimental Settings}

\subsection{Datasets}

\subsubsection{Binarized UltraFeedback} 
We train models with Binarized UltraFeedback Dataset \cite{ultrafeedback}. It consists of 64K preference pairs from diverse resources. 

\subsubsection{Llama3 UltraFeedback\footnote{https://huggingface.co/datasets/princeton-nlp/
llama3-ultrafeedback.}} 
Because there is a possibility that Binarized Ultrafeedback data was used in the training phase of Llama3-8B-instruct, \cite{simpo} proposed new dataset. The data is created base on responses generated by Llama3-8B-Instruct by using the Binarized Ultrafeedback prompts. Among these responses, the highest scoring response and the lowest scoring response, which are scored by reward model \cite{pairrm}, are used to form preference pairs. In this study, models trained using this dataset is labeled with the \textbf{v0.1} tag.

\subsection{Evaluation}
The trained models are evaluated on AlpacaEval2.0 and MT-Bench.

\subsubsection{Alpaca Eval 2.0} 

Alpaca Eval 2.0 \cite{alpaca_eval, alpaca_eval_lc} consists of 805 prompts. The responses generated using these prompts are compared against those produced by GPT-4-Turbo. Through this comparison, Alpaca Eval 2.0 quantify the model's performance by calculating the percentage of instances where its response surpass those of GPT-4-Turbo, expressed as a win rate (\textbf{WR}). AlpacaEval 2.0 also provides length controlled win rate (\textbf{LC}) that considers bias due to length.

% from each of these prompts are compared with the answer responses generated by using gpt-4-turbo, and the performance of the model is measured by the win rate (\textbf{WR}). AlpacaEval 2.0 also provides length controlled win rate (\textbf{LC}) that consider bias due to length.

\subsubsection{MT-Bench}

MT-Bench \cite{mtbench} is a multi-turn benchmark consisting of 80 distinct instructions to evaluate model performance. Model generated responses from these prompts are scored by using GPT-4. The benchmark's strength lies in its diverse category coverage, enabling comprehensive model assessment across multiple dimensions.

\subsection{Models and Baselines}
To compare across different model families, we use Mistral-7B-Base \cite{mistral} and Llama3-8B-Instruct \cite{llama3} as base model for preference optimization. 
We compare MIPO with DPO and also with SimPO, which utilizes average log likelihood. Additionally, results are compared with offline preference optimization methods, such as ORPO and KTO.

We implement MIPO, DPO and SimPO by using TRL \cite{trl} and the alignment book \cite{Tunstall_The_Alignment_Handbook}. 
When the Alpaca Eval 2.0 scores for models trained with DPO and SimPO are lower than those reported in the reference\footnote{\label{simpo-git}https://github.com/princeton-nlp/SimPO}, we adapts the reference values for a fair comparison. For MT-Bench evaluations, we utilize the checkpoints in reference to generate responses and evaluate. Additionally, we reference results from it for ORPO and KTO.

\section{Result and Analysis}

\subsection {Benchmark Results}
\label{benchmark_results}

As shown in Table \ref{table:overall}, MIPO consistently achieves higher scores compared to DPO and demonstrates outstanding performance relative to other methods in the most cases.

Comparative analysis using Alpaca Eval 2.0 reveals that MIPO consistently and significantly outperforms DPO across all experiments. Moreover, MIPO achieves performance levels comparable to SimPO, which had previously demonstrated the highest performance.

In MT-Bench, MIPO consistently exhibits enhanced performance relative to DPO and SimPO across all experiments.

\begin{figure}[t]
  \centering
  \includegraphics[scale=0.35]{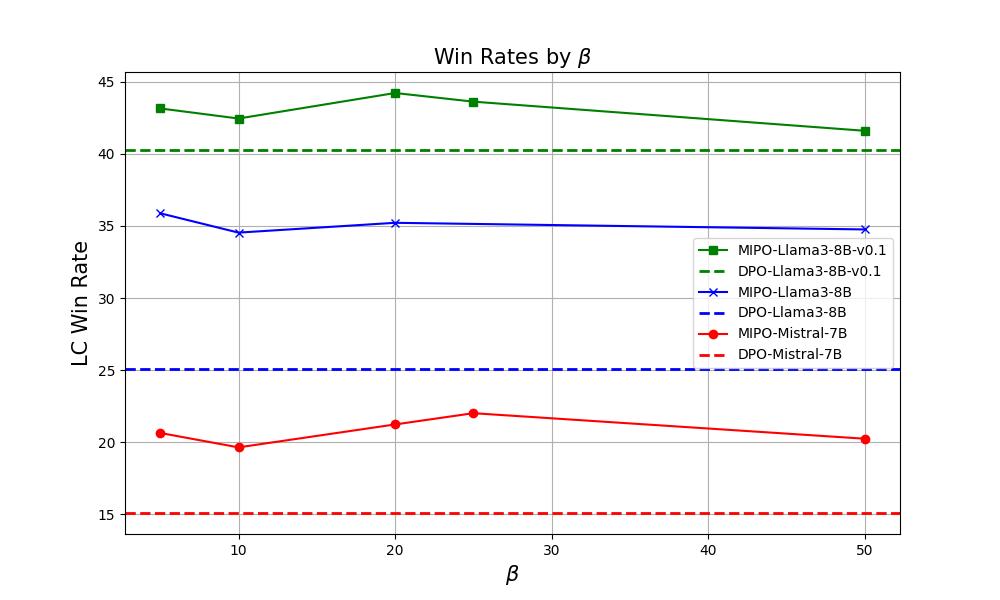}
  \caption{Alpaca Eval 2.0 scores in Mistral and Llama3 based on $\beta$. The dotted line represents the performance of DPO.}
  \label{fig:beta-performance}
\end{figure}

\subsection{Performance Based on $\beta$}
\label{6.2:training_stablity}

One of the advantages of MIPO is the ease of hyperparameter tuning. MIPO objective contains only a single hyperparameter, $\beta$, allowing for optimal model training by adjusting just this one. Figure \ref{fig:beta-performance} illustrates how the model's performance varies with different $\beta$ in Mistral-7B and Llama-8B. As depicted in Figure \ref{fig:beta-performance}, MIPO maintains exceptionally high performance across a similar beta range ($[5, 50]$), demonstrating robustness across various models and datasets. The optimal model configuration is consistently identified within this range. 

In conclusion, MIPO demonstrates a significant advantage: it consistently produces models that substantially outperform DPO and approach optimal performance levels, achieved through the tuning of a single hyperparameter, $\beta$, within a moderate range. This capability persists across diverse model architectures and datasets, underscoring MIPO's robustness and effectiveness.

\subsection {Analysis about Average Log Likelihood}

\begin{figure}[t]
  \centering
  \includegraphics[scale=0.4]{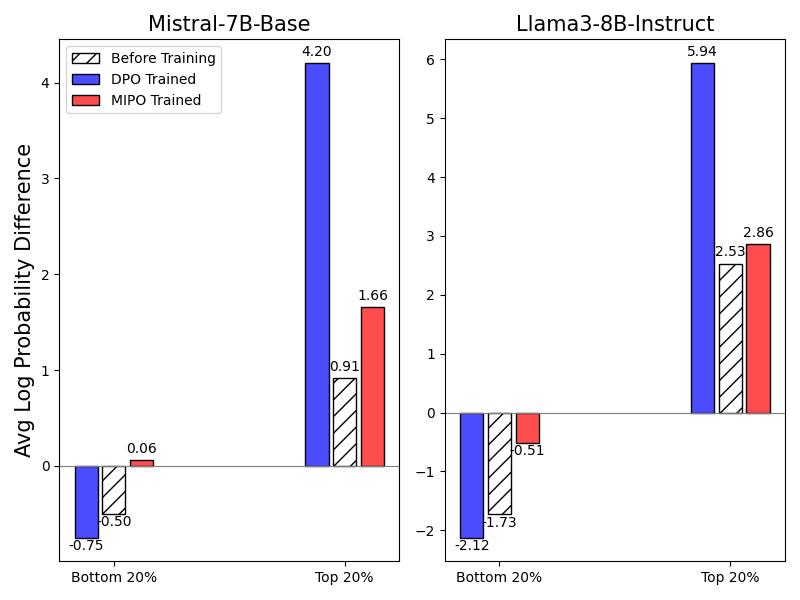}
  \caption{The difference in average log likelihood changes after training for both MIPO and DPO, as applied to Mistral-7B-Base and Llama3-8B-Instruct.}
  \label{avg log prob analysis}
\end{figure}

Figure \ref{avg log prob analysis}, represents the average log likelihood difference between chosen and rejected responses for the model on the evaluation dataset, showing how this difference changes after training. It specifically highlights how the values for instances in the top 20\% and bottom 20\% of average log likelihood differences in reference model have evolved. 

At this point, the top 20\% are instances with a large average log likelihood difference in reference model, indicating they are already well-aligned data, while the bottom 20\% are poorly aligned and require more training. 
% The results for the overall distribution can be found in Appendix \ref{appendix:avg}.

In the bottom 20\%, the average log likelihood difference for DPO actually decrease, whereas for MIPO, the average log likelihood clearly increase. Conversely, in the top 20\%, the average log likelihood for DPO increase significantly, while for MIPO, it only increase slightly. This pattern is observed in both the Llama3-8B and Mistral-7B.

This indicates that in DPO, the data that is already well-aligned continued to be better aligned through further training, while the data that is not well-aligned do not see significant improvement. 
%This confirms the earlier point made in Section \$\ref{3.3} that DPO loss does not adequately reflect the reference model's degree of alignment.
However, in MIPO, the training is operated to maintain performance on well-aligned data while significantly improving the alignment of poorly aligned data. MIPO achieves the intended outcome described in Section \$\ref{4.2}, thereby effectively enhancing model alignment.

\subsection{Analysis about MIPO objective function} \label{loss analysis}

\begin{figure}[t]
    \begin{subfigure}[t]{0.5\textwidth}
        \includegraphics[width=\textwidth]{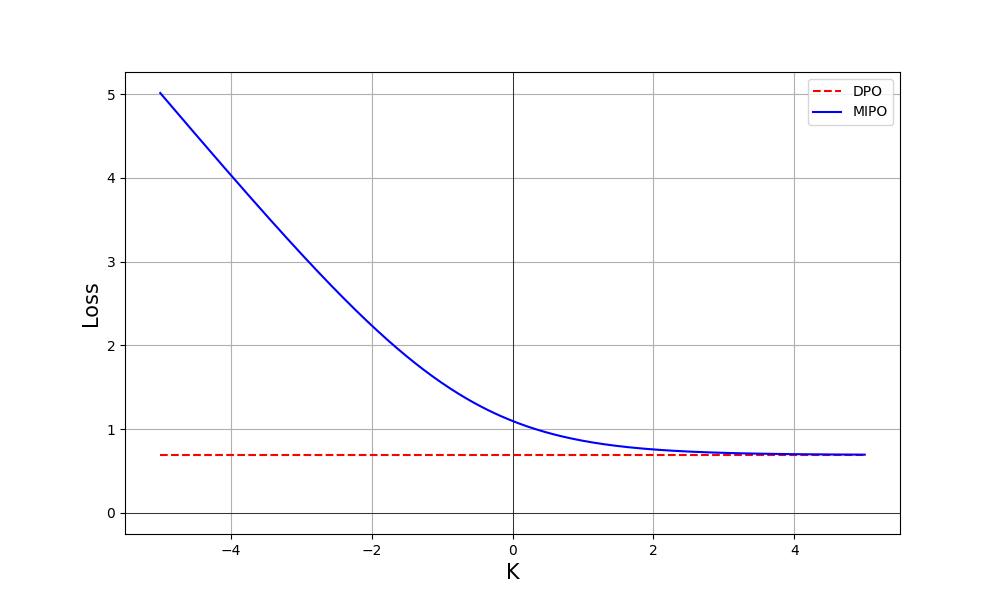}
        \caption{MIPO loss in early stages of training}
        \label{fig:loss_graph(b)}
    \hfill
    \end{subfigure}

    \begin{subfigure}[t]{0.5\textwidth}
        \includegraphics[width=\textwidth]{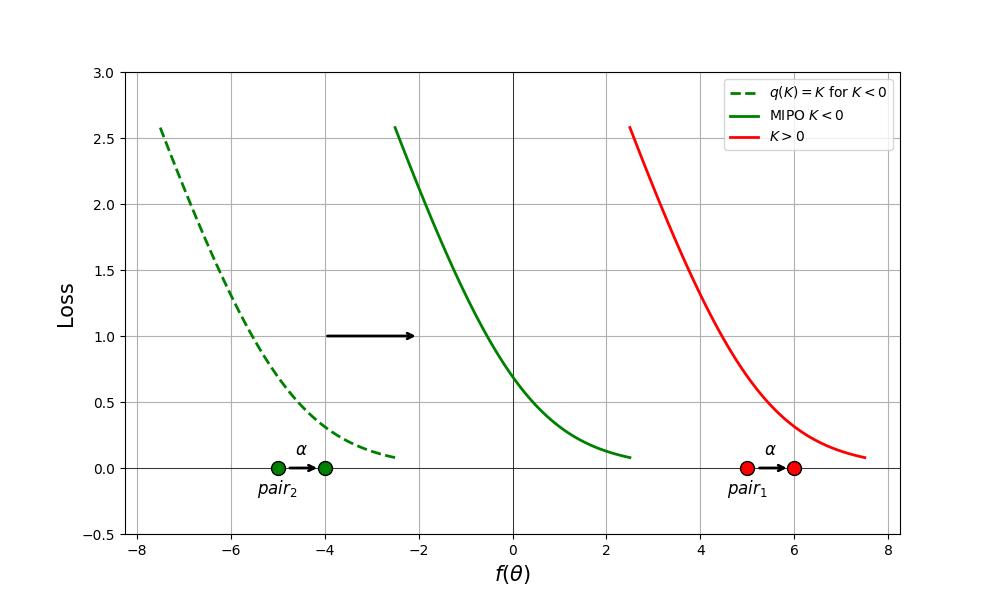}
        \caption{MIPO loss in high $K$ and low $K$}
        \label{fig:loss_graph(c)}
    \hfill
    \end{subfigure}
    
\end{figure}

As seen in Eq \eqref{eqn:5}, the MIPO objective can be expressed as the difference between the average log likelihood of the chosen response and rejected response in policy model and minus $q(K)$ consists of values calculated from the reference model.

% The reason for setting $q(K)$ to $ln(1+e^K)$ is because the function converges to K as K increases and to 0 as K decreases for adjusting the MIPO objective appropriately for high K value and low K value. 

Let's examine how the MIPO objective behaves during the training process in two scenarios.
\subsubsection{Early Stage in Training}

In the early stage of training, there is minimal difference between the reference model and the policy model. Therefore, the average log likelihood difference of the policy model does not significantly diverge from that of the reference model ($\pi_{ref} \approx \pi_{\theta}$). Consequently, the MIPO loss can be written as $\log\left(1+\ e^{-\left(K-\ln\left(1+e^{K}\right)\right)}\right)$. However, DPO loss for all instance initially 0. This can be observed in Figure \ref{fig:loss_graph(b)}.

\subsubsection{Loss Reflection During Training}

Next, let's examine how the loss for two pairs, $pair_1$ which has high $K$ value and $pair_2$ which has low $K$ value, behave during training. Suppose that the average log likelihood difference of the policy model, $f(\theta)$, increases by $\alpha > 0$ compared to the reference model for both pairs ($f(\theta) = K + \alpha$). 

In Figure \ref{fig:loss_graph(c)}, the red section represents $pair_1$. Since $pair_1$ has a high $K$, the MIPO objective is expressed as $-log sigmoid(f(\theta)-K)$ (the red line in the figure). Therefore, the MIPO loss is $-log sigmoid (\alpha)$, as we can be seen in the graph. Next, $pair_2$ is represented by the green section. Since $K$ is low, the MIPO objective is expressed as $-log sigmoid(f(\theta))$ (the green line in the figure). Therefore, the MIPO loss is $-log sigmoid(K+\alpha)$, which is larger than the loss for $pair_1$. Thus, even with the same amount of increase, $pair_2$ has a larger loss, indicating that training is accelerated for pairs with lower $K$.

Additionally, the figure's dotted line facilitates a comparative analysis between the scenarios where the $q(K)$ is simply $K$ and 0. In dotted line, even if $K$ is low, the loss is calculated based on the $K$. Thus, when the same increase occurs, the loss is calculated equally for both high and low $K$ pairs, causing the model to train with the same weight for both pairs.

As a result, the MIPO objective results in a relatively large loss when 
$K$ is low, indicating data that is poorly aligned. Thus, more extensive training can occur on poorly aligned data. Conversely, in the case of well-aligned data, the intervention from the reference model is substantial, causing the objective to be calculated based on the values of the reference model. This prevents the policy model from diverging significantly from the reference model.

\section{Conclusion}
In DPO, rewards are calculated based on the reference model for all pair data without considering how well the reference model is aligned with the given pair data. Therefore, DPO does not distinguish between instances that require more training ant those that only need minimal training. 
In this paper, we proposed \textbf{Modulated Intervention Preference Optimization (MIPO)} as a method to address and improve upon this issue. MIPO adjusts the objective based on the degree of alignment of the reference model on the given instances. 
For pairs that require more learning, MIPO reduces the intervention of the reference model, allowing the policy model to diverge from it and find better weights. Conversely, for pairs that are better aligned, the intervention of the reference model is maintained, ensuring that the policy model does not significantly diverge from the reference model.

Through experiments, we found that models trained using MIPO demonstrated significantly improved performance compared to those trained using DPO. Moreover, we observed a notable increase in the average log likelihood difference for instances with initially small differences from the reference model, aligning with our expectations compared to DPO.

\textbf{Limitations \& Future Work}
While MIPO achieved significantly better performance than DPO, there are still several areas for improvement and further investigation

\subsubsection{Average log likelihood is not an absolute measure of the degree of alignment }

The degree of preference between the chosen and rejected responses can vary for each preference pair. In some cases, the chosen and rejected responses might be decided by a very subtle difference, while in others, the difference could be significant. If a given preference pair has only a slight difference, the model may be well-aligned, but the average log probability difference ($K$) is unlikely to be large. Therefore, it is difficult to accurately assert that a large $K$ indicates superiority on a particular preference pair. The $K$ alone does not provide an absolute measure of performance across different preference pairs.

Although MIPO does not account for the difficulty differences between preference pairs, it is likely that pairs where the model was poorly aligned improved more, as higher average log likelihoods typically indicate better performance for each pair.

\subsubsection{Experiments on Various Adaptation Terms}
While we used $ln(1 + e^K)$ to construct the loss for MIPO, there are various functions that could be used as adaptation term. To achieve the same effect as MIPO, the function $q$ should use an alternative function that converges to $K$ when $K$ is large and to 0 when $K$ is small. However, experiments exploring all possible functions for this have not yet been conducted.

% \section{Acknowledgments}

\bibliography{aaai25}

\begin{thebibliography}{37}
\providecommand{\natexlab}[1]{#1}

\bibitem[{AI@Meta(2024)}]{llama3}
AI@Meta. 2024.
\newblock Llama 3 Model Card.

\bibitem[{Askell et~al.(2021)Askell, Bai, Chen, Drain, Ganguli, Henighan, Jones, Joseph, Mann, DasSarma et~al.}]{rlhf-ref2}
Askell, A.; Bai, Y.; Chen, A.; Drain, D.; Ganguli, D.; Henighan, T.; Jones, A.; Joseph, N.; Mann, B.; DasSarma, N.; et~al. 2021.
\newblock A general language assistant as a laboratory for alignment.
\newblock \emph{arXiv preprint arXiv:2112.00861}.

\bibitem[{Azar et~al.(2024)Azar, Guo, Piot, Munos, Rowland, Valko, and Calandriello}]{ipo}
Azar, M.~G.; Guo, Z.~D.; Piot, B.; Munos, R.; Rowland, M.; Valko, M.; and Calandriello, D. 2024.
\newblock A general theoretical paradigm to understand learning from human preferences.
\newblock In \emph{International Conference on Artificial Intelligence and Statistics}, 4447--4455. PMLR.

\bibitem[{Bai et~al.(2022{\natexlab{a}})Bai, Jones, Ndousse, Askell, Chen, DasSarma, Drain, Fort, Ganguli, Henighan et~al.}]{safety}
Bai, Y.; Jones, A.; Ndousse, K.; Askell, A.; Chen, A.; DasSarma, N.; Drain, D.; Fort, S.; Ganguli, D.; Henighan, T.; et~al. 2022{\natexlab{a}}.
\newblock Training a helpful and harmless assistant with reinforcement learning from human feedback.
\newblock \emph{arXiv preprint arXiv:2204.05862}.

\bibitem[{Bai et~al.(2022{\natexlab{b}})Bai, Kadavath, Kundu, Askell, Kernion, Jones, Chen, Goldie, Mirhoseini, McKinnon et~al.}]{safety2}
Bai, Y.; Kadavath, S.; Kundu, S.; Askell, A.; Kernion, J.; Jones, A.; Chen, A.; Goldie, A.; Mirhoseini, A.; McKinnon, C.; et~al. 2022{\natexlab{b}}.
\newblock Constitutional ai: Harmlessness from ai feedback.
\newblock \emph{arXiv preprint arXiv:2212.08073}.

\bibitem[{Bradley and Terry(1952)}]{bt}
Bradley, R.~A.; and Terry, M.~E. 1952.
\newblock Rank analysis of incomplete block designs: I. The method of paired comparisons.
\newblock \emph{Biometrika}, 39(3/4): 324--345.

\bibitem[{Brown et~al.(2020)Brown, Mann, Ryder, Subbiah, Kaplan, Dhariwal, Neelakantan, Shyam, Sastry, Askell et~al.}]{plm3}
Brown, T.; Mann, B.; Ryder, N.; Subbiah, M.; Kaplan, J.~D.; Dhariwal, P.; Neelakantan, A.; Shyam, P.; Sastry, G.; Askell, A.; et~al. 2020.
\newblock Language models are few-shot learners.
\newblock \emph{Advances in neural information processing systems}, 33: 1877--1901.

\bibitem[{Casper et~al.(2023)Casper, Davies, Shi, Gilbert, Scheurer, Rando, Freedman, Korbak, Lindner, Freire et~al.}]{rlhf-problem1}
Casper, S.; Davies, X.; Shi, C.; Gilbert, T.~K.; Scheurer, J.; Rando, J.; Freedman, R.; Korbak, T.; Lindner, D.; Freire, P.; et~al. 2023.
\newblock Open problems and fundamental limitations of reinforcement learning from human feedback.
\newblock \emph{arXiv preprint arXiv:2307.15217}.

\bibitem[{Chowdhery et~al.(2023)Chowdhery, Narang, Devlin, Bosma, Mishra, Roberts, Barham, Chung, Sutton, Gehrmann et~al.}]{plm}
Chowdhery, A.; Narang, S.; Devlin, J.; Bosma, M.; Mishra, G.; Roberts, A.; Barham, P.; Chung, H.~W.; Sutton, C.; Gehrmann, S.; et~al. 2023.
\newblock Palm: Scaling language modeling with pathways.
\newblock \emph{Journal of Machine Learning Research}, 24(240): 1--113.

\bibitem[{Christiano et~al.(2017)Christiano, Leike, Brown, Martic, Legg, and Amodei}]{rlhf-ref1}
Christiano, P.~F.; Leike, J.; Brown, T.; Martic, M.; Legg, S.; and Amodei, D. 2017.
\newblock Deep reinforcement learning from human preferences.
\newblock \emph{Advances in neural information processing systems}, 30.

\bibitem[{Chung et~al.(2024)Chung, Hou, Longpre, Zoph, Tay, Fedus, Li, Wang, Dehghani, Brahma et~al.}]{instruction-tuning}
Chung, H.~W.; Hou, L.; Longpre, S.; Zoph, B.; Tay, Y.; Fedus, W.; Li, Y.; Wang, X.; Dehghani, M.; Brahma, S.; et~al. 2024.
\newblock Scaling instruction-finetuned language models.
\newblock \emph{Journal of Machine Learning Research}, 25(70): 1--53.

\bibitem[{Cui et~al.(2023)Cui, Yuan, Ding, Yao, Zhu, Ni, Xie, Liu, and Sun}]{ultrafeedback}
Cui, G.; Yuan, L.; Ding, N.; Yao, G.; Zhu, W.; Ni, Y.; Xie, G.; Liu, Z.; and Sun, M. 2023.
\newblock UltraFeedback: Boosting Language Models with High-quality Feedback.
\newblock arXiv:2310.01377.

\bibitem[{Dai et~al.(2023)Dai, Pan, Sun, Ji, Xu, Liu, Wang, and Yang}]{rlhf-safe}
Dai, J.; Pan, X.; Sun, R.; Ji, J.; Xu, X.; Liu, M.; Wang, Y.; and Yang, Y. 2023.
\newblock Safe rlhf: Safe reinforcement learning from human feedback.
\newblock \emph{arXiv preprint arXiv:2310.12773}.

\bibitem[{Dubois et~al.(2024)Dubois, Galambosi, Liang, and Hashimoto}]{alpaca_eval_lc}
Dubois, Y.; Galambosi, B.; Liang, P.; and Hashimoto, T.~B. 2024.
\newblock Length-Controlled AlpacaEval: A Simple Way to Debias Automatic Evaluators.
\newblock \emph{arXiv preprint arXiv:2404.04475}.

\bibitem[{Ethayarajh et~al.(2024)Ethayarajh, Xu, Muennighoff, Jurafsky, and Kiela}]{kto}
Ethayarajh, K.; Xu, W.; Muennighoff, N.; Jurafsky, D.; and Kiela, D. 2024.
\newblock Kto: Model alignment as prospect theoretic optimization.
\newblock \emph{arXiv preprint arXiv:2402.01306}.

\bibitem[{Gorbatovski et~al.(2024)Gorbatovski, Shaposhnikov, Malakhov, Surnachev, Aksenov, Maksimov, Balagansky, and Gavrilov}]{dpo-probelm-2}
Gorbatovski, A.; Shaposhnikov, B.; Malakhov, A.; Surnachev, N.; Aksenov, Y.; Maksimov, I.; Balagansky, N.; and Gavrilov, D. 2024.
\newblock Learn your reference model for real good alignment.
\newblock \emph{arXiv preprint arXiv:2404.09656}.

\bibitem[{Havrilla et~al.(2024)Havrilla, Du, Raparthy, Nalmpantis, Dwivedi-Yu, Zhuravinskyi, Hambro, Sukhbaatar, and Raileanu}]{rlhf-helpful2}
Havrilla, A.; Du, Y.; Raparthy, S.~C.; Nalmpantis, C.; Dwivedi-Yu, J.; Zhuravinskyi, M.; Hambro, E.; Sukhbaatar, S.; and Raileanu, R. 2024.
\newblock Teaching large language models to reason with reinforcement learning.
\newblock \emph{arXiv preprint arXiv:2403.04642}.

\bibitem[{Hong, Lee, and Thorne(2024)}]{orpo}
Hong, J.; Lee, N.; and Thorne, J. 2024.
\newblock Orpo: Monolithic preference optimization without reference model.
\newblock \emph{arXiv preprint arXiv:2403.07691}, 2(4): 5.

\bibitem[{Jiang et~al.(2023)Jiang, Sablayrolles, Mensch, Bamford, Chaplot, Casas, Bressand, Lengyel, Lample, Saulnier et~al.}]{mistral}
Jiang, A.~Q.; Sablayrolles, A.; Mensch, A.; Bamford, C.; Chaplot, D.~S.; Casas, D. d.~l.; Bressand, F.; Lengyel, G.; Lample, G.; Saulnier, L.; et~al. 2023.
\newblock Mistral 7B.
\newblock \emph{arXiv preprint arXiv:2310.06825}.

\bibitem[{Jiang, Ren, and Lin(2023)}]{pairrm}
Jiang, D.; Ren, X.; and Lin, B.~Y. 2023.
\newblock {LLM-Blender}: Ensembling Large Language Models with Pairwise Comparison and Generative Fusion.
\newblock In \emph{Proceedings of the 61th Annual Meeting of the Association for Computational Linguistics (ACL 2023)}.

\bibitem[{Korbak et~al.(2023)Korbak, Shi, Chen, Bhalerao, Buckley, Phang, Bowman, and Perez}]{rlhf-helpful}
Korbak, T.; Shi, K.; Chen, A.; Bhalerao, R.~V.; Buckley, C.; Phang, J.; Bowman, S.~R.; and Perez, E. 2023.
\newblock Pretraining language models with human preferences.
\newblock In \emph{International Conference on Machine Learning}, 17506--17533. PMLR.

\bibitem[{Li et~al.(2023)Li, Zhang, Dubois, Taori, Gulrajani, Guestrin, Liang, and Hashimoto}]{alpaca_eval}
Li, X.; Zhang, T.; Dubois, Y.; Taori, R.; Gulrajani, I.; Guestrin, C.; Liang, P.; and Hashimoto, T.~B. 2023.
\newblock AlpacaEval: An Automatic Evaluator of Instruction-following Models.
\newblock \url{https://github.com/tatsu-lab/alpaca_eval}.

\bibitem[{Liu, Liu, and Cohan(2024)}]{dpo-problem}
Liu, Y.; Liu, P.; and Cohan, A. 2024.
\newblock Understanding Reference Policies in Direct Preference Optimization.
\newblock \emph{arXiv preprint arXiv:2407.13709}.

\bibitem[{Meng, Xia, and Chen(2024)}]{simpo}
Meng, Y.; Xia, M.; and Chen, D. 2024.
\newblock Simpo: Simple preference optimization with a reference-free reward.
\newblock \emph{arXiv preprint arXiv:2405.14734}.

\bibitem[{Ouyang et~al.(2022)Ouyang, Wu, Jiang, Almeida, Wainwright, Mishkin, Zhang, Agarwal, Slama, Ray et~al.}]{rlhf-ref3}
Ouyang, L.; Wu, J.; Jiang, X.; Almeida, D.; Wainwright, C.; Mishkin, P.; Zhang, C.; Agarwal, S.; Slama, K.; Ray, A.; et~al. 2022.
\newblock Training language models to follow instructions with human feedback.
\newblock \emph{Advances in neural information processing systems}, 35: 27730--27744.

\bibitem[{Peng et~al.(2023)Peng, Song, Tian, Jin, Mi, and Yu}]{rlhf-problem2}
Peng, B.; Song, L.; Tian, Y.; Jin, L.; Mi, H.; and Yu, D. 2023.
\newblock Stabilizing RLHF through advantage model and selective rehearsal.
\newblock \emph{arXiv preprint arXiv:2309.10202}.

\bibitem[{Rafailov et~al.(2024)Rafailov, Sharma, Mitchell, Manning, Ermon, and Finn}]{dpo}
Rafailov, R.; Sharma, A.; Mitchell, E.; Manning, C.~D.; Ermon, S.; and Finn, C. 2024.
\newblock Direct preference optimization: Your language model is secretly a reward model.
\newblock \emph{Advances in Neural Information Processing Systems}, 36.

\bibitem[{Ramamurthy et~al.(2022)Ramamurthy, Ammanabrolu, Brantley, Hessel, Sifa, Bauckhage, Hajishirzi, and Choi}]{instruction-tuning2}
Ramamurthy, R.; Ammanabrolu, P.; Brantley, K.; Hessel, J.; Sifa, R.; Bauckhage, C.; Hajishirzi, H.; and Choi, Y. 2022.
\newblock Is reinforcement learning (not) for natural language processing: Benchmarks, baselines, and building blocks for natural language policy optimization.
\newblock \emph{arXiv preprint arXiv:2210.01241}.

\bibitem[{Schulman et~al.(2017)Schulman, Wolski, Dhariwal, Radford, and Klimov}]{ppo}
Schulman, J.; Wolski, F.; Dhariwal, P.; Radford, A.; and Klimov, O. 2017.
\newblock Proximal policy optimization algorithms.
\newblock \emph{arXiv preprint arXiv:1707.06347}.

\bibitem[{Thoppilan et~al.(2022)Thoppilan, De~Freitas, Hall, Shazeer, Kulshreshtha, Cheng, Jin, Bos, Baker, Du et~al.}]{instruction-tuning3}
Thoppilan, R.; De~Freitas, D.; Hall, J.; Shazeer, N.; Kulshreshtha, A.; Cheng, H.-T.; Jin, A.; Bos, T.; Baker, L.; Du, Y.; et~al. 2022.
\newblock Lamda: Language models for dialog applications.
\newblock \emph{arXiv preprint arXiv:2201.08239}.

\bibitem[{Touvron et~al.(2023)Touvron, Lavril, Izacard, Martinet, Lachaux, Lacroix, Rozi{\`e}re, Goyal, Hambro, Azhar et~al.}]{plm2}
Touvron, H.; Lavril, T.; Izacard, G.; Martinet, X.; Lachaux, M.-A.; Lacroix, T.; Rozi{\`e}re, B.; Goyal, N.; Hambro, E.; Azhar, F.; et~al. 2023.
\newblock Llama: Open and efficient foundation language models.
\newblock \emph{arXiv preprint arXiv:2302.13971}.

\bibitem[{Tunstall et~al.()Tunstall, Beeching, Lambert, Rajani, Huang, Rasul, Bartolome, M.~Rush, and Wolf}]{Tunstall_The_Alignment_Handbook}
Tunstall, L.; Beeching, E.; Lambert, N.; Rajani, N.; Huang, S.; Rasul, K.; Bartolome, A.; M.~Rush, A.; and Wolf, T. ????
\newblock {The Alignment Handbook}.

\bibitem[{von Werra et~al.(2020)von Werra, Belkada, Tunstall, Beeching, Thrush, Lambert, and Huang}]{trl}
von Werra, L.; Belkada, Y.; Tunstall, L.; Beeching, E.; Thrush, T.; Lambert, N.; and Huang, S. 2020.
\newblock TRL: Transformer Reinforcement Learning.
\newblock \url{https://github.com/huggingface/trl}.

\bibitem[{Wang et~al.(2024)Wang, Zheng, Chen, Liu, Dou, Huang, Shen, Jin, Zhou, Shi et~al.}]{rlhf-problem3}
Wang, B.; Zheng, R.; Chen, L.; Liu, Y.; Dou, S.; Huang, C.; Shen, W.; Jin, S.; Zhou, E.; Shi, C.; et~al. 2024.
\newblock Secrets of rlhf in large language models part ii: Reward modeling.
\newblock \emph{arXiv preprint arXiv:2401.06080}.

\bibitem[{Wei et~al.(2021)Wei, Bosma, Zhao, Guu, Yu, Lester, Du, Dai, and Le}]{instruction-tuning4}
Wei, J.; Bosma, M.; Zhao, V.~Y.; Guu, K.; Yu, A.~W.; Lester, B.; Du, N.; Dai, A.~M.; and Le, Q.~V. 2021.
\newblock Finetuned language models are zero-shot learners.
\newblock \emph{arXiv preprint arXiv:2109.01652}.

\bibitem[{Xu et~al.(2024)Xu, Sharaf, Chen, Tan, Shen, Van~Durme, Murray, and Kim}]{cpo}
Xu, H.; Sharaf, A.; Chen, Y.; Tan, W.; Shen, L.; Van~Durme, B.; Murray, K.; and Kim, Y.~J. 2024.
\newblock Contrastive preference optimization: Pushing the boundaries of llm performance in machine translation.
\newblock \emph{arXiv preprint arXiv:2401.08417}.

\bibitem[{Zheng et~al.(2023)Zheng, Chiang, Sheng, Zhuang, Wu, Zhuang, Lin, Li, Li, Xing, Zhang, Gonzalez, and Stoica}]{mtbench}
Zheng, L.; Chiang, W.-L.; Sheng, Y.; Zhuang, S.; Wu, Z.; Zhuang, Y.; Lin, Z.; Li, Z.; Li, D.; Xing, E.~P.; Zhang, H.; Gonzalez, J.~E.; and Stoica, I. 2023.
\newblock Judging LLM-as-a-judge with MT-Bench and Chatbot Arena.
\newblock arXiv:2306.05685.

\end{thebibliography}

\end{document}